%% file: main.tex
\documentclass[10pt,twocolumn,letterpaper]{article}

\usepackage{cvpr}              
\input{preamble}
\definecolor{cvprblue}{rgb}{0.21,0.49,0.74}
\usepackage[pagebackref,breaklinks,colorlinks,allcolors=cvprblue]{hyperref}
\usepackage{algorithm}
\usepackage{algorithmic}
\def\paperID{} 
\def\confName{CVPR}
\def\confYear{2026}

\title{Virtual Nodes Guided Dynamic Graph Neural Network for Brain Tumor Segmentation with Missing Modalities}

\author{
Sha Tao \qquad Jiao Pan \qquad Yu Guo \qquad Chao Yao\thanks{Corresponding author}\\
University of Science and Technology Beijing, Beijing, China\\
}

\begin{document}
\setcounter{footnote}{1}
\maketitle
\input{sec/0_abstract}    
\input{sec/1_intro}
{
    \small
    \bibliographystyle{ieeenat_fullname}
    \bibliography{main}
}


\end{document}

%% file: sec/0_abstract.tex
\begin{abstract}
Multimodal magnetic resonance imaging (MRI) is crucial for brain tumor segmentation, with many methods leveraging its four key modalities to capture complementary information for effective sub-region analysis. However, the absence of several modalities is very common in practice, leading to severe performance degradation in existing full-modality segmentation methods. Limited by the structured data model, recent works often adopt a multi-stage training strategy for full-modality and missing-modality scenarios, which increases training costs and inadequately addresses the interference of miss. In this work, we propose a graph-based one-stage framework for robust brain tumor segmentation with missing modalities. Specifically, we introduce modality-specific virtual nodes that serve as supplementary information sources to compensate for missing modalities. To enhance model robustness against arbitrary modality combinations, we leverage the inherent flexibility of graph networks to devise a dynamic connection strategy. This mechanism dynamically adjusts the adjacency matrix based on modality availability, preserving beneficial information flow while mitigating interference effects caused by missing modalities. Furthermore, we introduce heterogeneous weight matrices to improve the graph network’s adaptability to multimodal scenarios. Extensive experiments on the BRATS-2018 and BRATS-2020 datasets demonstrate that our method outperforms the state-of-the-art methods on almost all subsets of incomplete modalities.
\end{abstract}

%% file: sec/1_intro.tex
\section{Introduction}
\label{sec:intro}

\begin{figure}
    \centering
    \includegraphics[width=8cm,trim=0 0 0 0,clip]{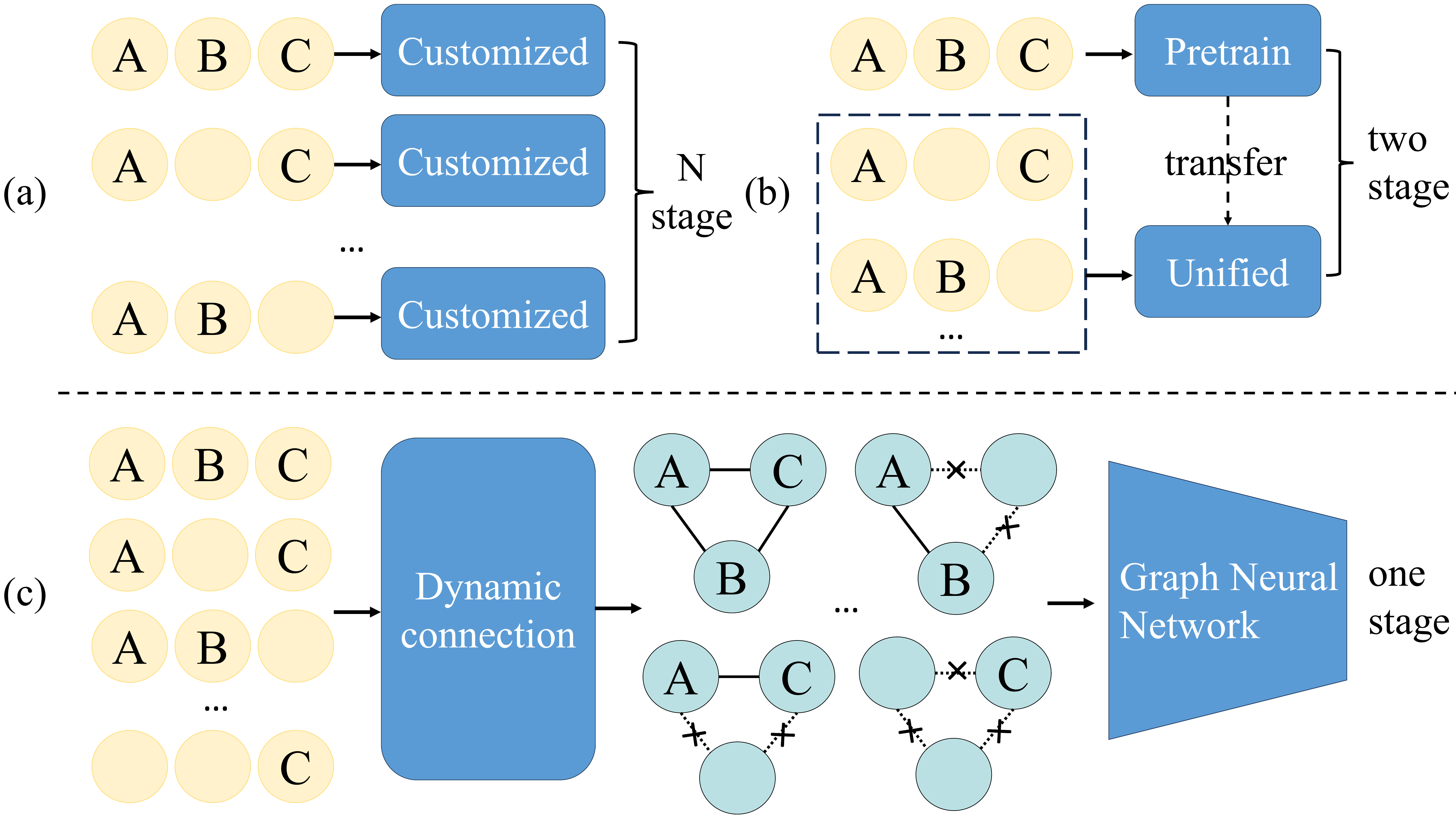}
    \caption{Comparison between our method and recent approaches. (a) Customized methods. (b) Recent unified methods. (c) Our graph-based approach.}\label{fig:f0}
\end{figure}
Multi-parametric magnetic resonance imaging (MRI) is the current standard for the quantitative evaluation of brain tumors in clinical practice~\cite{c:1}. Specifically, four MRI modalities are commonly used: T1-weighted (T1), contrast-enhanced T1-weighted (T1c), T2-weighted (T2), and Fluid Attenuated Inversion Recovery (FLAIR) images. Each modality delivers specific and valuable features that collectively improve the precision of brain tumor segmentation. In recent years, plenty of works have explored the use of deep learning methods for brain tumor segmentation~\cite{c:2,c:3}. These methods were optimized for the full-modality scenario and had already achieved excellent results. However, it is not always possible to acquire a complete set of MRIs due to practical constraints, such as image corruption, acquisition protocols, or unsuitable conditions of patients, and in more practical settings, some modalities may even be unavailable during training~\cite{re:0}. In such cases, these advanced methods often suffer from severe performance degradation.

To address the challenge of missing modalities, numerous methods have been proposed. A straightforward solution involves training a customized model for each possible subset of modalities~\cite{c:4,c:5}, as shown in Fig.~\ref{fig:f0} (a). While this strategy can yield promising results, it suffers from prohibitively high training and deployment costs, which grow exponentially with the number of modalities. This fundamental limitation severely restricts their applicability in real-world clinical settings. As a result, recent research has shifted toward learning a unified model capable of handling incomplete multimodal inputs, primarily by employing random modality dropout during training ~\cite{c:6,c:7}. However, structured data models (e.g., CNNs or Transformers) heavily rely on complete contextual information and struggle to accommodate all modality combinations within a single-stage training framework. To overcome this limitation, many existing methods adopt a two-stage training strategy: either employing knowledge distillation~\cite{c:8,a:5,c:10,cvpr:2}, where a full-modality model guides the learning of incomplete inputs, or training a generative model to reconstruct the missing modalities~\cite{c:11,c:12}. 

However, these two-stage solutions ultimately stem from the rigidity of structured architectures such as CNNs and Transformers, whose computation paths are predefined for a fixed set of modalities and thus rely on complete cross-modality correspondence. As shown in Fig.~\ref{fig:f0}(a–b), this fixed connectivity makes both customized and unified structured models sensitive to missing-modality patterns, since the expected feature interactions become ill-defined once one or more modalities are absent. In contrast, a graph formulation operates on the actually observed modality set: each modality is treated as a node, and missing modalities simply correspond to absent nodes or edges. As illustrated in Fig.~\ref{fig:f0}(c), such adaptive connectivity enables message passing only among available modalities, making the model more robust under incomplete multimodal settings.

In this work, we propose a plug-and-play graph-based one-stage learning framework. Specifically, features extracted from modality-specific encoders are treated as nodes, which are then processed by a Graph Attention Network (GAT) for feature representation and fusion~\cite{a:1}. To enhance the GAT’s suitability for multimodal learning, we introduce a heterogeneous weight matrix that allows for modality-aware attention. To mitigate information loss due to missing modalities, we incorporate zero-initialized virtual nodes to capture modality-invariant features, thereby providing supplementary information even in the total absence of certain modalities. Furthermore, for the overall multimodal graph composed of both basic and virtual nodes, we design an adaptive edge connectivity strategy that supports diverse modality combinations, where information flows unidirectionally from existing nodes to those representing missing modalities. Experimenting on the public BraTS-2018 and BraTS-2020 datasets demonstrates that our method outperforms the state-of-the-art methods on almost all subsets of incomplete modalities. Even in the most extreme single-modality scenarios, our method still maintains robust performance.

Our key contributions can be summarized as follows:
\begin{itemize}
\item We propose a plug-and-play graph-based framework to handle missing modalities in brain tumor segmentation, which can be easily integrated into existing systems. The unstructured nature of graph architectures enables one-stage training directly on incomplete data, eliminating the need for additional supervision under full-modality settings.
\item We introduce zero-initialized virtual nodes to mitigate information loss caused by missing modalities. By enabling collaborative representation between base and virtual nodes, our model preserves comprehensive contextual information across arbitrary modality combinations.
\item We design a dynamic edge connection mechanism where information flows unidirectionally from existing nodes to those representing missing modalities. This allows the model to flexibly adapt to varying combinations of available modalities within a single unified architecture.
\end{itemize}

\begin{figure*}[t]
\centering
    \includegraphics[width= \linewidth]{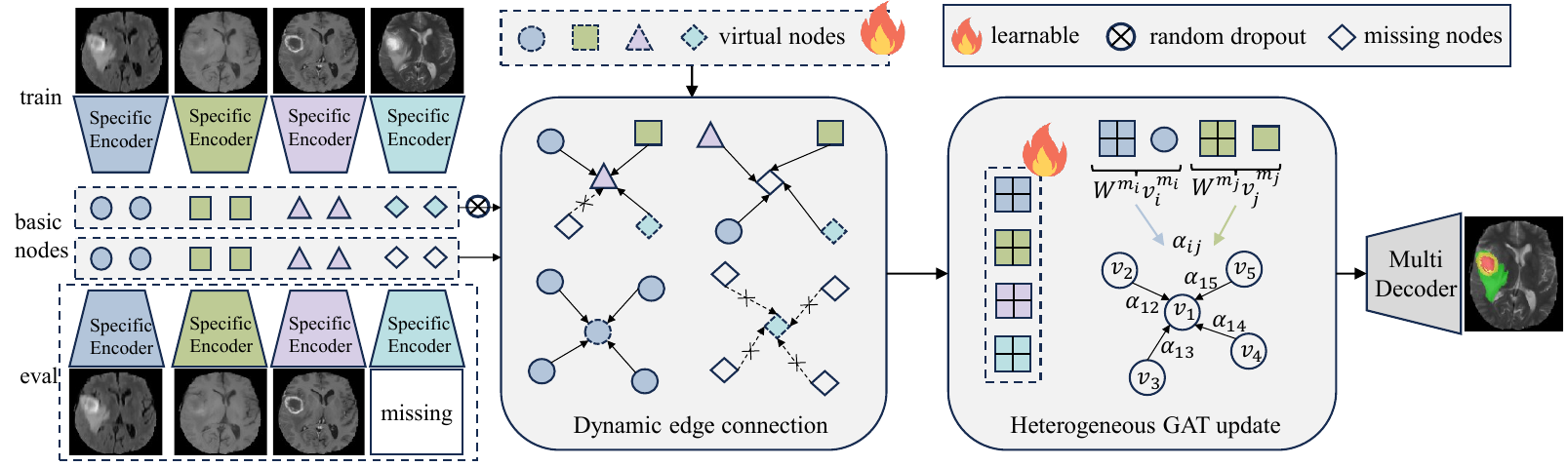}
    \caption{Overview of the proposed framework. The detailed dynamic edge connection is shown in Fig. \ref{fig:f2} and Algorithm \ref{alg:algorithm}.}\label{fig:overview}
\end{figure*}

\section{Related Work}
\subsection{Multimodal Brain Tumor Segmentation with Missing Modalities}Existing methods for brain tumor segmentation with missing modalities can be broadly categorized into explicit and implicit approaches according to how missing modalities are represented.

Explicit (reconstruction-based) methods recover missing modality information either at the image level or the feature level. Image-level methods synthesize missing modalities for full-modality segmentation (e.g., using GANs)~\cite{a:2,a:3,tmi:1,tmi:2,a:4}, yet accurately reconstructing critical missing modalities from the remaining ones is often difficult or even infeasible. Feature-level methods instead infer representative embeddings for missing modalities. For example, Shen and Gao~\cite{c:15} used adversarial learning to align feature distributions, while Wang et al.~\cite{c:11} proposed a Shared-Specific Feature Modeling framework for modality-invariant and modality-specific representations. Although such methods promote cross-modality alignment, they may suppress modality-specific cues, which is detrimental when essential modality information is absent. In addition, both image- and feature-level approaches usually require auxiliary generative networks, increasing model complexity and computational cost.

Implicit methods focus on learning a multimodal latent feature space. Hetero-modal variational encoder-decoder~\cite{c:13} incorporated multimodal variational auto-encoders to reconstruct modalities from a shared latent variable, while LS3M~\cite{cvpr:1} employs learnable sorting and state space modeling for end-to-end segmentation under arbitrary modality combinations. Many methods also leverage knowledge distillation~\cite{a:7}, where teacher (full-modality) and student (missing-modality) models are trained to transfer latent knowledge~\cite{c:8,a:5,c:10,cvpr:2}. Compared with explicit reconstruction, implicit methods are easier to train, but often provide limited compensation for missing modalities and still rely on two-stage training.

Overall, most existing methods still rely on structured architectures and often require two-stage optimization to cope with missing modalities. In contrast, our method adopts a graph-based formulation that combines explicit missing-modality representation with implicit feature learning.

\subsection{Graph Neural Networks for Medical Image Segmentation}Graph Neural Networks (GNNs) ~\cite{c:16} have previously been applied to medical image segmentation tasks. Yan et al.~\cite{c:17} first used the SLIC algorithm \cite{a:11} to cluster MRIs into supervoxels, and then predicted the tissue type of each supervoxel. Inspired by this work, subsequent research has generally followed a similar workflow~\cite{c:18,c:19,a:12}. However, such methods typically depend on superpixel-based representations, which hinder the preservation of modality-specific characteristics and reduce the effectiveness of cross-modal information fusion. These limitations are especially critical in scenarios with missing modalities.

In addition, several prior works have incorporated graph-inspired mechanisms to tackle missing modalities. Yang et al.~\cite{tmi:1} introduced a graph-attention fusion module for multimodal MR image synthesis and tumor segmentation, and Zhao et al.~\cite{tmi:4} proposed a modality-adaptive feature interaction framework based on graph representation concepts. However, these methods mainly incorporate graph-inspired operations into otherwise structured architectures, and do not explicitly address how missing modalities should be represented or how information should propagate under different missing-modality patterns. By contrast, our work formulates missing-modality interaction through a plug-and-play graph interface, where virtual nodes are introduced to account for absent modalities and the graph connectivity is dynamically adapted to the observed modality subset. As such, the proposed framework differs from prior graph-inspired methods not only in architectural form, but also in how modality absence is modeled and how cross-modality information flow is defined.

\section{Method}
An overview of the proposed framework is shown in Fig.~\ref{fig:overview}. We first extract modality-specific features from the multimodal MRI inputs using dedicated encoders. In parallel, we construct a graph containing both image-derived nodes and zero-initialized virtual nodes, together with a default adjacency matrix defined for the complete-modality setting. The extracted features are then mapped to their corresponding nodes and processed by a Graph Attention Network (GAT) equipped with a heterogeneous weight matrix to better model multimodal relationships. To support arbitrary combinations of available modalities, we adopt random modality dropout, replacing missing-modality nodes with zero vectors and dynamically updating graph connectivity via our adaptive edge strategy. Finally, the enriched node representations are fed into the decoder to produce the segmentation map.

The main difference between training and inference lies in the availability of modalities. During training, all modalities are accessible, and missing-modality scenarios are simulated via random modality dropout. Below, we will provide a detailed explanation of each module.

\subsection{Node Construction}We denote the complete set of modalities by $M=\{\text{FLAIR}, \text{T1c}, \text{T1}, \text{T2}\}$. Given an input image of $\mathbf{x}^{m}\in \mathbb{R}^{1\times D\times H\times W} $, where $W$, $H$, $D$ are the width, height, depth of the image, and $m \in M$, we utilize the specific convolutional encoder to generate feature maps. In this condition, the original images of all modalities are available, so we can get independent and complete features of each modality. Assume that the features obtained through the encoder are denoted as $\mathbf{f}^{m}\in \mathbb{R}^{C\times D^{'}\times H^{'}\times W^{'}} $, where $C, D^{'}, H^{'}$ and $ W^{'} $ refer to the number of channels and the corresponding feature dimensions obtained after several convolutional downsampling operations. Through the transformation of feature dimensions, we convert them into nodes $\mathbf{v}^{m}\in \mathbb{R}^{C\times F} $, where $F = D^{'}\times H^{'}\times W^{'} $. 

To mitigate the information loss introduced by missing modalities, we introduce a set of learnable, zero-initialized virtual nodes $\mathbf{p}^{m}\in \mathbb{R}^{C_p\times F} $ for each modality. These nodes act as modality-agnostic anchors that interact with the available modality nodes through message passing, allowing the model to preserve a stable and coherent representation even when some modalities are absent.

Finally, the missing-aware virtual nodes are attached to the basic nodes $\mathbf{v}^{m}$ along with the input-length dimension to form extended nodes $\mathbf{v}^{m}_p$:
\begin{equation}
\mathbf{v}_p^{m}=\left[\mathbf{v}^{m} ; \mathbf{p}^{m}\right]
\end{equation}
where $[\cdot ; \cdot]$ represents the concatenation operation, $\mathbf{v}_p^{m}\in \mathbb{R}^{P\times F}$, and $P=C+C_p$. For a complete multimodal graph network, there are a total of $4\times P$ nodes. We refer to the $i$-th node of the $m$-th modality as $\mathbf{v}_i^{m}$, and if the modality is not specifically indicated, $m$ can be omitted.

\begin{figure}
    \centering
    \includegraphics[width=.78\columnwidth,trim=0 0 0 0,clip]{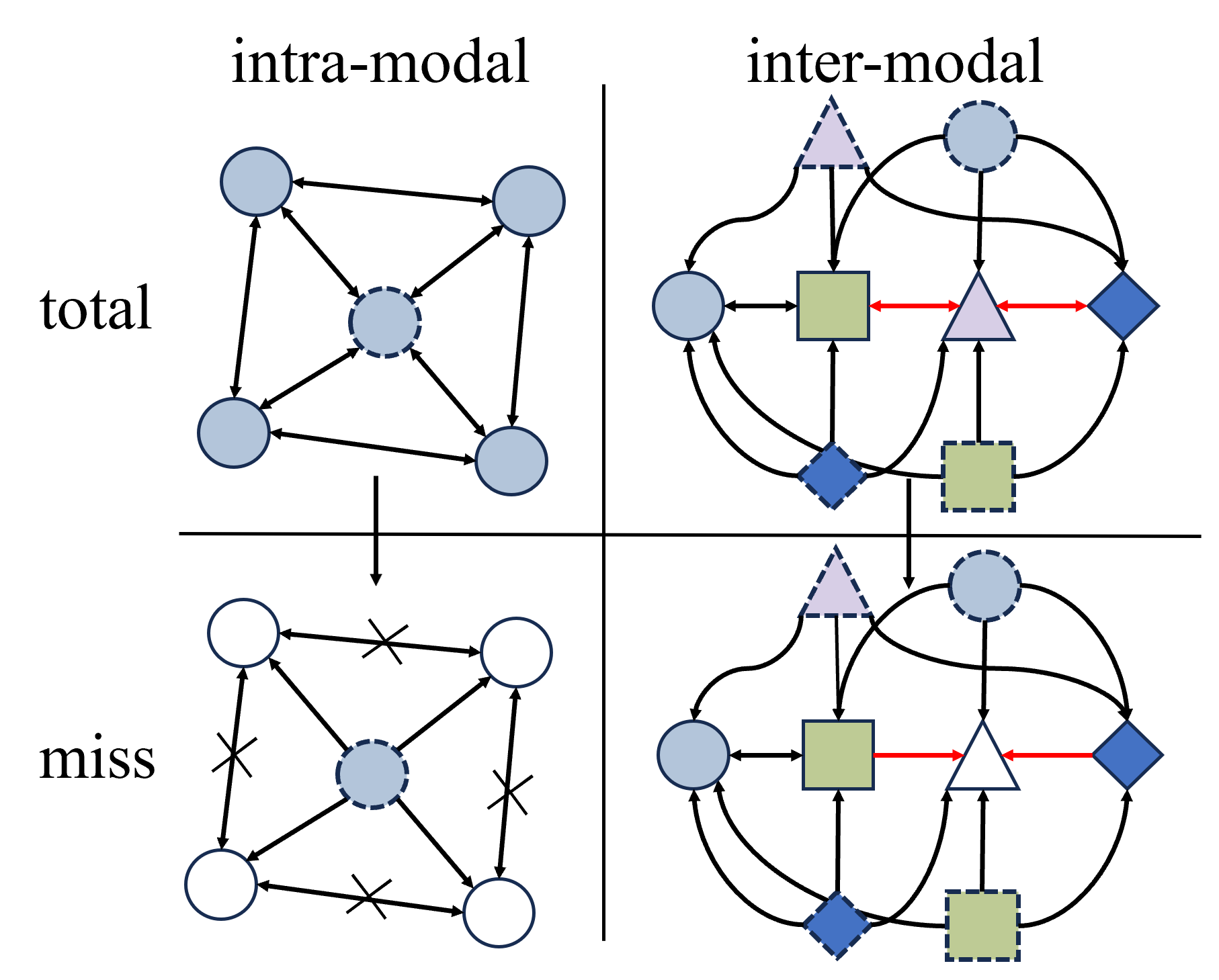}
    \caption{Dynamic edge connection, illustrating the simulation from the full-modality scenario to the missing. The dashed shapes represent virtual nodes; the hollow shapes represent missing nodes; and '×' denotes a broken connection.}\label{fig:f2}
\end{figure}

\subsection{Edge Construction and Dynamic Adaptability}Predefined or fixed connectivity becomes unreliable when modalities are missing, as absent inputs lead to noisy or invalid edges. Therefore, we construct dynamic connections that adjust the graph topology according to the available modalities. In this section, we detail our multimodal edge design. Since the Graph Attention Network (GAT) achieves weighted connections between different nodes through an attention mechanism, the construction of edge structures primarily serves to describe the connectivity between nodes with value 0 and 1. We use $e_{ij}$ to represent an edge from $\mathbf{v}_j$ to $\mathbf{v}_i$.

As shown in upper part of Fig. \ref{fig:f2}, the multimodal graph structure in a full-modal scenario constructs \( e_{ij} \) according to the following rules: intra-modal nodes are bidirectionally connected, and so are corresponding nodes across different modalities. Virtual nodes are updated only by nodes of their own modality but participate in updating all other nodes.

During training, we randomly apply dropout to mask certain modality nodes, which dynamically alters the edge connections. Without loss of generality, suppose modality \( m \) is dropped (in practice, the number of dropped modalities is randomly selected between 0 and \( N-1 \), where \( N \) is the total number of modalities). As shown in lower part of Fig. \ref{fig:f2}, for any basic node \( i \) within modality \( m \), its connections to all other nodes \( j \) are cut, meaning \( e_{ji} = 0 \) while \( e_{ij} \) remains unchanged. This approach allows us to reconstruct the missing modality features by utilizing other modality features and the invariant features of the the missing modality itself, while avoiding noise caused by reconstruction from interfering with other existing modalities. The detailed procedure is presented in Algorithm \ref{alg:algorithm}.

Fundamentally, our approach shares the same objective as customized methods—defining a specific model for each possible modality combination. However, the key distinction lies in our novel use of graph networks' structural flexibility. Through dynamic connectivity, we implicitly encode N distinct model configurations within a single unified graph architecture. This enables our model to generalize across arbitrary modality combinations at inference time, achieving both adaptability and efficiency within a single-stage framework.
\begin{algorithm}[tb]
	\caption{Dynamic Edge Connection Method}
	\begin{algorithmic}[1]
		\STATE \textbf{Input:}
		\STATE \quad Two nodes $\mathbf{v}_i^m$ and $\mathbf{v}_j^n$, where $m$ and $n$ represent their 
                    respective modality; basic nodes set $\mathcal{N}_b$; virtual nodes set
                    $\mathcal{N}_v$; total nodes set $\mathcal{N}_t$; random missing node $\mathbf{v}_k$.
   
		\STATE \textbf{Initial:}
		\STATE \quad \textbf{if} $m == n$ \textbf{then}
            \STATE \quad \quad $e_{ij} = e_{ji} = 1$
            \STATE \quad \textbf{else}
            \STATE \quad \quad \textbf{if} $\mathbf{v}_i^m \in \mathcal{N}_b$ and $\mathbf{v}_j^n \in \mathcal{N}_b$ \textbf{then}
            \STATE \quad \quad \quad \textbf{if} $i = j$ \textbf{then}
            \STATE \quad \quad \quad \quad $e_{ij} = e_{ji} = 1$
            \STATE \quad \quad \quad \textbf{else}
            \STATE \quad \quad \quad \quad $e_{ij} = e_{ji} = 0$
            \STATE \quad \quad \quad \textbf{end if}
            \STATE \quad \quad \textbf{else if} $\mathbf{v}_i^m \in \mathcal{N}_b$ and $\mathbf{v}_j^n \in \mathcal{N}_v$ \textbf{then}
            \STATE \quad \quad \quad \quad $e_{ij} = 1; e_{ji} = 0$
            \STATE \quad \quad \textbf{else if} $\mathbf{v}_i^m \in \mathcal{N}_v$ and $\mathbf{v}_j^n \in \mathcal{N}_b$ \textbf{then}
            \STATE \quad \quad \quad \quad $e_{ij} = 0; e_{ji} = 1$
            \STATE \quad \quad \textbf{else}
            \STATE \quad \quad \quad \quad $e_{ij} = e_{ji} = 0$
            \STATE \quad \quad \textbf{end if}
            \STATE \quad \textbf{end if}
            \STATE \textbf{Dynamic change:}
            \STATE \quad \textbf{for} $\mathbf{v}_i \in \mathcal{N}_t$ \textbf{do}
            \STATE \quad \quad $e_{ik} = 0$
            \STATE \quad \textbf{end}
\end{algorithmic}
\label{alg:algorithm}
\end{algorithm} 

\subsection{Heterogeneous Graph Attention Network} To calculate the weights when updating nodes, Graph Attention Network (GAT) performs self-attention on the nodes~\cite{a:1}. Since GAT was originally designed for single-modality data, it did not account for the differences between modalities in a multimodal scenario, resulting in the weight matrix $\mathbf{W}$ being shared across all nodes. As shown in Fig. \ref{fig:overview}, in this work, we heterogenize the weight matrix $\mathbf{W}$ to better accommodate the needs of multimodal features:
\begin{equation}
\boldsymbol{\beta}_{ij}=a(\mathbf{W}^{m}\mathbf{v}_i^{m},\mathbf{W}^{n}\mathbf{v}_j^{n})
\end{equation}
where ${m, n} \in \{FLAIR, T1c, T1, T2\}$, $a$ is an attentional mechanism and $ \boldsymbol{\beta}_{ij}$ represents the attention coefficients between node $\mathbf{v}_i$ and $\mathbf{v}_j$. To simulate the modality-missing scenario, during training, we apply a random dropout to mask the basic nodes of 0 to \( N-1 \) modalities by zero-filling them. Meanwhile, the edge structure is updated according to the rules mentioned earlier. When updating node \( \mathbf{v}_i \), we calculate the relevance coefficient \( \boldsymbol{\beta}_{ij} \) only for nodes \( \mathbf{v}_j \) where \( e_{ij} = 1 \), and get the weight coefficient of \( \mathbf{v}_j \) relative to \( \mathbf{v}_i \):
\begin{equation}
\boldsymbol{\alpha}_{ij}=\mathrm{softmax}_j(\boldsymbol{\beta}_{ij})=\frac{\exp(\boldsymbol{\beta}_{ij})}{\sum_{k\in\mathcal{N}_i}\exp(\boldsymbol{\beta}_{ik})}
\end{equation}
where $\mathcal{N}_i$ is the neighborhood of node i in the graph, i.e. $\mathcal{N}_{i}\:=\:\{j\:|\:e_{ij}\:=\:1\}$. In practice, to handle the disconnected edges caused by missing modalities  (i.e., \( e_{ij} = 0 \)) and ensure gradient propagation, we did not simply remove them from the computation graph. Instead, we employed a softer approach by assigning a very small value to the corresponding attention weights. This approach reduces the interference from the missing modalities while maintaining differentiability.

In GAT's experiments, the attention mechanism $a$ is a single-layer feedforward neural network, parametrized by a weight vector $\mathbf{a}$, which is also shared between nodes. We heterogenized the $\mathbf{a}$ as well, and the final weight coefficient can be expressed as:
\begin{equation}
    \boldsymbol{\alpha}_{ij}=\frac{\exp\left(\mathrm{LeakyReLU}\left(\mathbf{a}^m[\mathbf{W}^{m}\mathbf{v}_i^{m}\|\mathbf{W}^{n}\mathbf{v}_j^{n}]\right)\right)}{\sum_{k\in\mathcal{N}_i}\exp\left(\mathrm{LeakyReLU}\left(\mathbf{a}^m[\mathbf{W}^{m}\mathbf{v}_i^{m}\|\mathbf{W}^{m_k}\mathbf{v}_k^{m_k}]\right)\right)}
\end{equation}

Subsequently, similar to GAT, a multi-head attention mechanism is introduced to update the nodes:
\begin{equation}
    \mathbf{v}_i'=\sigma\left(\frac{1}{K}\sum_{k=1}^K\sum_{j\in\mathcal{N}_i}\boldsymbol{\alpha}_{ij}^k\mathbf{W}_k^m\mathbf{v}_j^{m}\right)
\end{equation}
where K is the number of independent attention mechanisms.

\begin{table*}[t]
\setlength{\tabcolsep}{.7mm}
\centering
\begin{tabular}{ccccc|cccccc|cccccc|ccccc}
\hline
\multicolumn{4}{c}{Modality} &  & \multicolumn{5}{c}{Enhancing tumor} &  & \multicolumn{5}{c}{Tumor core} &  & \multicolumn{5}{c}{Whole tumor}                  \\ \hline
Fl    & T1        & T1c       & T2        &  & HVED & mmF &   M\textsuperscript{3}AE & MMC & Ours &  & HVED & mmF & M\textsuperscript{3}AE & MMC  & Ours &  & HVED  & mmf &     M\textsuperscript{3}AE & MMC   & Ours  \\ \hline
$\bullet$   & $\circ$   & $\circ$   & $\circ$ &  
& 23.8 & 39.3 & 35.6 & \textbf{49.6} & 46.0 &                    
& 57.9 & 61.2 & 66.4 & 70.0 & \textbf{74.9} &  
& 84.4 & 86.1 & 88.7 & 86.2 & \textbf{89.0} \\
$\circ$   & $\bullet$   & $\circ$ & $\circ$   &  
& 8.6  & 32.5 & 37.1 & 42.9 & \textbf{43.9} &                    
& 33.9 & 56.6 & 66.1 & 69.3 & \textbf{72.0} &  
& 49.5 & 67.5 & 74.4 & 78.6 & \textbf{78.9} \\
$\circ$   & $\circ$ & $\bullet$   & $\circ$   &  
& 57.6 & 72.6 & 73.7 & \textbf{79.0} & 78.0 &       	 	              
& 59.6 & 75.4 & 82.9 & \textbf{86.6} & 84.1 &  
& 53.6 & 72.2 & 75.8 & \textbf{80.4} & 76.6 \\
$\circ$ & $\circ$   & $\circ$   & $\bullet$   &  
& 22.8 & 43.1 & 47.6 & \textbf{50.7} & 50.0 &                 	 	     
& 54.7 & 64.2 & 69.4 & 69.7 & \textbf{73.6} &  
& 79.8 & 81.2 & 84.8 & 84.1 & \textbf{84.9} \\
$\bullet$   & $\bullet$   & $\circ$ & $\circ$ &  
& 28.0 & 43.0 & 41.2 & 48.1 & \textbf{50.3} &                    
& 61.1 & 65.9 & 70.8 & 72.0 & \textbf{77.7} &  
& 85.7 & 87.1 & 89.0 & 87.7 & \textbf{90.0} \\
$\bullet$   & $\circ$ & $\bullet$ & $\circ$   &  
& 68.4 & 75.1 & 75.0 & 79.5 & \textbf{81.0} &     	 	
& 75.1 & 77.9 & 84.4 & 86.7 & \textbf{87.1} &  
& 85.9 & 87.3 & 89.7 & 88.7 & \textbf{90.2} \\
$\bullet$ & $\circ$ & $\circ$   & $\bullet$   &  
& 32.3 & 47.5 & 45.4 & 48.3 & \textbf{53.4} &      	 	                
& 62.7 & 69.8 & 70.9 & 72.4 & \textbf{76.4} &  
& 87.6 & 87.6 & 89.9 & 88.0 & \textbf{90.0} \\
$\circ$   & $\bullet$ & $\bullet$   & $\circ$ &  
& 61.1 & 74.0 & 74.7 & 78.7 & \textbf{80.0} &            	         
& 67.6 & 78.6 & 83.4 & \textbf{86.6} & 86.2 &  
& 64.2 & 74.4 & 77.2 & 80.2 & \textbf{80.5} \\
$\circ$ & $\bullet$   & $\circ$   & $\bullet$ &  
& 24.3 & 45.0 & 48.7 & 46.9 & \textbf{50.4} &                    
& 56.3 & 69.4 & 71.8 & 74.0 & \textbf{76.0} &   	 	
& 81.6 & 82.2 & 86.7 & 85.9 & \textbf{86.5} \\
$\circ$ & $\circ$   & $\bullet$ & $\bullet$   &  
& 67.8 & 74.5 & 75.3 & 79.1 & \textbf{80.2} &      	 	              
& 73.9 & 78.6 & 84.2 & 86.7 & \textbf{87.0} &  
& 81.3 & 83.0 & 86.3 & \textbf{86.3} & 86.2 \\
$\bullet$ & $\bullet$ & $\bullet$ & $\circ$   &  
& 68.6 & 75.5 & 74.0 & 78.1 & \textbf{81.3} &                	 	     
& 77.1 & 79.8 & 84.1 & 86.6 & \textbf{87.3} &  
& 86.7 & 87.3 & 88.9 & 88.3 & \textbf{90.3} \\
$\bullet$ & $\bullet$ & $\circ$   & $\bullet$ &   	 	
& 32.3 & 47.7 & 44.8 & 50.7 & \textbf{54.5} &                    
& 63.1 & 71.5 & 72.7 & 67.8 & \textbf{77.6} &  
& 88.1 & 87.8 & 89.9 & 88.5 & \textbf{90.2} \\
$\bullet$ & $\circ$   & $\bullet$ & $\bullet$ &  
& 68.9 & 75.7 & 73.8 & 79.1 & \textbf{80.7} &                    
& 76.8 & 79.6 & 84.6 & 86.6 & \textbf{87.4} &   	 	 
& 88.1 & 88.1 & 90.2 & 88.9 & \textbf{90.3} \\
$\circ$   & $\bullet$ & $\bullet$ & $\bullet$ &  		
& 67.8 & 74.8 & 75.4 & 79.2 & \textbf{80.3} &                    
& 75.3 & 80.4 & 84.4 & 86.7 & \textbf{87.0} &  
& 82.3 & 82.7 & 85.7 & \textbf{86.8} & 86.5 \\
$\bullet$ & $\bullet$ & $\bullet$ & $\bullet$ &  
& 69.0 & 77.6 & 75.5 & 80.1 & \textbf{80.8} &             		       
& 77.7 & 85.8 & 84.5 & \textbf{87.4} & 87.3 &  
& 88.5 & 89.6 & 90.1 & 89.0 & \textbf{90.3} \\
\hline 	 	
\multicolumn{4}{c}{Mean}                      &  
& 46.8 & 59.9 & 59.9 & 64.7 & \textbf{66.0} &                    
& 64.8 & 73.0 & 77.4 & 79.3 & \textbf{81.4} &  
& 79.2 & 82.9 & 85.8 & 85.8 & \textbf{86.7} \\
\hline
\end{tabular}
\caption{Brain tumor segmentation results for different modality combinations on BraTS 2018 dataset. Present and missing modalities are denoted by $\bullet$ and $\circ$. The \textbf{best} results are bolded.
}
\label{tab:brats18}
\end{table*}

\subsection{Overall Optimization:} Due to the inconsistency between the data distribution perceived by the specific encoder and the subsequent model, and to ensure the quality of the generated features, we introduced an additional specific decoder for optimization. Additionally, for the complete features updated by the GAT, a multimodal decoder is employed to perform segmentation prediction. To optimize virtual nodes, which represent modality-invariant features, some methods minimize the differences between modalities~\cite{c:11}. However, this manual intervention tends to reduce the richness of the generated information, with only shared features generated across modalities. Therefore, we adopt a self-learning approach, allowing the model to autonomously discover and enhance the modality-invariant features without imposing rigid constraints. This method leverages the natural alignment of features across modalities, encouraging the model to learn richer and more diverse representations.

In general, our method generates two types of segmentation maps during training, including $\mathbf{y}^s$ of specific decoder and $\mathbf{y}^m$ of multimodal decoder. Therefore, the overall segmentation loss is defined between the ground-truth label $\mathbf{y}$ and the two estimated segmentation maps respectively:
\begin{equation}
    \mathcal{L}_{\mathrm{total~}}=\sum_{i\in M}\mathcal{L}(\mathbf{y}^s_{i}, \mathbf{y})+\mathcal{L}(\mathbf{y}^m, \mathbf{y})
\end{equation}
where $M=\{\text{FLAIR}, \text{T1c}, \text{T1}, \text{T2}\}$ and $\mathcal{L}$ is the Dice loss~\cite{c:20}. Note that the specific encoder is only used during the training phase, while during inference, the segmentation results are obtained solely from the multimodal decoder.

\section{Experiments}
\subsection{Datasets and Evaluation Metrics} We evaluate our method on two widely used brain tumor segmentation benchmarks: BraTS 2018 and 2020~\cite{a:13}. Both datasets provide multi-modal MRI scans for each subject, including FLAIR, T1, T1ce, and T2 sequences, along with expert-annotated tumor labels. The annotations consist of four categories: necrotic and non-enhancing tumor core (NET), enhancing tumor (ET), edema (ED), and background (BG). These categories are further grouped into three sub-regions for evaluation: (1) whole tumor (WT), including all tumor regions; (2) tumor core (TC), which comprises NET and ET; and (3) enhancing tumor (ET). The BraTS 2018 and 2020 datasets contain 285 and 369 annotated cases respectively. Following previous works, we split them into training/validation/testing subsets with ratios of 199/29/57 for BraTS 2018~\cite{c:13} and 219/50/100 for BraTS 2020~\cite{c:21, c:22}. Following the BraTS challenge, we use the Dice similarity coefficient (DSC) for performance quantification.

\subsection{Implementation Details} We adopt 3D U-Net~\cite{c:7} as the backbone, using its encoder as modality-specific feature extractors and reusing the decoder for both single-modality and multimodal decoding. The key difference lies in the number of channels: one for the specific decoder and four for the multimodal decoder. The framework is implemented in PyTorch 1.7 and trained on four NVIDIA 4090 GPUs with a batch size of 1 for 1000 epochs. We use the Adam optimizer~\cite{a:14} with an initial learning rate of 0.0002 and a cosine decay schedule~\cite{a:15}. During training, input volumes are randomly cropped from 240×240×155 to 128×128×128 voxels to reduce memory consumption. We preserve complete modality inputs for the modality-specific encoders, and simulate missing modalities by randomly zeroing out selected features when constructing graph nodes. At inference time, all $2^N - 1$ modality combinations are evaluated by skipping the corresponding encoders and graph connections for missing modalities. For fair and consistent evaluation on full-resolution images, we adopt a sliding window strategy following~\cite{c:23}.

\subsection{Comparison with the State of the Art (SOTA)} The experimental results for brain tumor segmentation with missing modalities on BraTS2018 are shown in Tab \ref{tab:brats18}, which compares our method with current state-of-the-art  methods, including U-HVED~\cite{c:13}, mmFormer~\cite{c:7}, M\textsuperscript{3}AE~\cite{c:12}, MMCFormer~\cite{c:24}. Among them, MMCFormer is customized and the others are unified. Notably, mmFormer serves as a representative one-stage method, whereas M\textsuperscript{3}AE exemplifies the two-stage paradigm.

\begin{table*}[t]
\setlength{\tabcolsep}{.7mm}
\centering
\begin{tabular}{ccccc|cccccc|cccccc|ccccc}
\hline
\multicolumn{4}{c}{Modality}                  &  & \multicolumn{5}{c}{Enhancing tumor}                      &  & \multicolumn{5}{c}{Tumor core}                       &  & \multicolumn{5}{c}{Whole tumor}                  \\ \hline
Fl    & T1        & T1c       & T2        &  & HVED & mmF &   M\textsuperscript{3}AE & LS3M & Ours &  & HVED & mmF & M\textsuperscript{3}AE & LS3M  & Ours &  & HVED  & mmF &     M\textsuperscript{3}AE & LS3M   & Ours  \\ \hline
$\bullet$   & $\circ$   & $\circ$   & $\circ$ &  
& 12.9 & 40.5 & 40.5 & 40.3 & \textbf{43.8} &                    
& 34.6 & 66.2 & 68.0 & \textbf{72.0} & 70.3 &  
& 69.9 & 82.4 & 86.5 & 88.7 & \textbf{89.0} \\
$\circ$   & $\bullet$   & $\circ$ & $\circ$   &  
& 7.3 & 34.0 & 39.9 & 36.6 & \textbf{44.4} &                    
& 27.3 & 61.2 & 66.0 & 66.7 & \textbf{71.6} &  
& 46.8 & 74.4 & 76.7 & 79.3 & \textbf{79.5} \\
$\circ$   & $\circ$ & $\bullet$   & $\circ$   &  
& 24.9 & 68.9 & 72.4 & 77.7 & \textbf{81.6} &       	 	              
& 35.5 & 78.0 & 81.4 & 83.6 & \textbf{86.9} &  
& 46.8 & 74.3 & 73.9 & 78.2 & \textbf{78.9} \\
$\circ$ & $\circ$   & $\circ$   & $\bullet$   &  
& 24.3 & 45.6 & 46.0 & 45.3 & \textbf{46.2} &                 	 	     
& 37.7 & 69.2 & 70.3 & \textbf{72.3} & 69.4 &  
& 54.0 & 83.1 & 86.1 & \textbf{87.3} & 85.1 \\
$\bullet$   & $\bullet$   & $\circ$ & $\circ$ &  
& 22.0 & 43.6 & 43.2 & 45.1 & \textbf{49.3} &                    
& 38.3 & 69.6 & 73.8 & 75.3 & \textbf{75.5} &  
& 58.3 & 84.6 & 89.4 & 90.6 & \textbf{90.7} \\
$\bullet$   & $\circ$ & $\bullet$ & $\circ$   &  
& 30.0 & 69.8 & 74.7 & 79.2 & \textbf{81.7} &     	 	
& 42.2 & 80.4 & 82.0 & 86.3 & \textbf{87.4} &  
& 61.5 & 84.5 & 89.5 & 91.0 & \textbf{91.7} \\
$\bullet$ & $\circ$ & $\circ$   & $\bullet$   &  
& 29.4 & 48.1 & 47.3 & \textbf{51.5} & 51.3 &      	 	                
& 43.4 & 71.6 & 75.0 & \textbf{76.9} & 74.7 &  
& 64.5 & 85.8 & 89.3 & 91.5 & \textbf{92.1} \\
$\circ$   & $\bullet$ & $\bullet$   & $\circ$ &  
& 33.6 & 71.1 & 75.4 & 79.5 & \textbf{79.6} &            	         
& 44.9 & 79.9 & 82.4 & \textbf{84.1} & 80.9 &  
& 62.9 & 78.0 & 78.1 & 82.6 & \textbf{87.5} \\
$\circ$ & $\bullet$   & $\circ$   & $\bullet$ &  
& 32.1 & 45.9 & 46.6 & 47.0 & \textbf{51.9} &                    
& 45.0 & 70.9 & 72.5 & 74.3 & \textbf{75.2} &   	 	
& 64.3 & 84.0 & 87.2 & \textbf{88.9} & 87.7 \\
$\circ$ & $\circ$   & $\bullet$ & $\bullet$   &  
& 36.2 & 70.7 & 76.8 & 78.7 & \textbf{81.8} &      	 	              
& 47.5 & 80.8 & 83.0 & 85.2 & \textbf{87.1} &  
& 65.8 & 84.1 & 87.4 & \textbf{89.2} & 87.7 \\
$\bullet$ & $\bullet$ & $\bullet$ & $\circ$   &  
& 39.4 & 70.1 & 75.9 & \textbf{80.8} & 80.0 &                	 	     
& 49.1 & 80.2 & 82.4 & 86.1 & \textbf{87.6} &  
& 67.0 & 85.3 & 90.0 & 91.5 & \textbf{91.6} \\
$\bullet$ & $\bullet$ & $\circ$   & $\bullet$ &   	 	
& 38.1 & 48.4 & 48.2 & 51.5 & \textbf{53.7} &                    
& 49.4 & 72.0 & 75.1 & \textbf{76.9} & 76.2 &  
& 68.4 & 86.2 & 90.4 & 91.5 & \textbf{91.7} \\
$\bullet$ & $\circ$   & $\bullet$ & $\bullet$ &  
& 40.9 & 71.6 & 77.1 & 80.5 & \textbf{81.5} &                    
& 51.3 & 81.3 & 83.1 & 85.8 & \textbf{87.4} &   	 	 
& 69.7 & 86.1 & 90.2 & 91.9 & \textbf{92.1} \\
$\circ$   & $\bullet$ & $\bullet$ & $\bullet$ &  		
& 43.2 & 70.7 & 77.4 & \textbf{81.5} & 80.2 &                    
& 52.7 & 81.1 & 84.1 & 85.8 & \textbf{87.4} &  
& 70.4 & 84.6 & 88.6 & \textbf{89.8} & 88.2 \\
$\bullet$ & $\bullet$ & $\bullet$ & $\bullet$ &  
& 45.3 & 71.4 & 78.0 & \textbf{80.9} & 79.8 &             		       
& 54.2 & 81.2 & 84.4 & 85.9 & \textbf{87.5} &  
& 71.4 & 86.4 & 90.6 & 92.2 & \textbf{92.3} \\
\hline 	 	
\multicolumn{4}{c}{Mean}                      &  
& 30.6 & 58.0 & 61.3 & 63.6 & \textbf{65.8} &                    
& 43.5 & 74.9 & 77.6 & 79.8 & \textbf{80.8} &  
& 62.8 & 82.9 & 86.3 & 88.2 & \textbf{88.4} \\
\hline
\end{tabular}
\caption{Brain tumor segmentation results for different modality combinations on BraTS2020 dataset. Present and missing modalities are denoted by $\bullet$ and $\circ$. The \textbf{best} results are bolded.
}
\label{tab:brats20}
\end{table*}

Compared with the latest state-of-the-art method MMCFormer, our approach achieves superior performance on BraTS 2018, with Dice improvements of 1.3\% on enhancing tumor (ET), 2.1\% on tumor core (TC), and 0.9\% on whole tumor (WT). While the average gains appear moderate, our method demonstrates consistent advantages across different modality combinations. Notably, when the most informative modality (T1ce) is missing (e.g., lines 5, 7, 9, and 12), we observe significant gains, with average improvements of 3.65\% on ET and 4.3\% on TC. This highlights the model's capacity to compensate for missing critical information. Overall, our method achieves the best Dice scores in 12 out of 15 combinations for ET, TC, and WT, respectively. In the remaining cases, the performance gap is clinically negligible, with average differences below 1.5\% compared to the top-performing baseline. Compared with other representative methods, our approach also achieves notable performance gains. It is worth noting that, compared with mmFormer, the primary difference lies in the mechanism for modeling multimodal interactions: mmFormer employs a Transformer between the encoder and decoder of a 3D U-Net, whereas our method adopts a graph neural network for the same purpose. Considering this key architectural distinction and the corresponding performance gains, the results suggest that graph-based modeling offers clear advantages for capturing modality relationships in missing-modality scenarios.
\begin{figure}
    \centering
    \includegraphics[width=\linewidth]{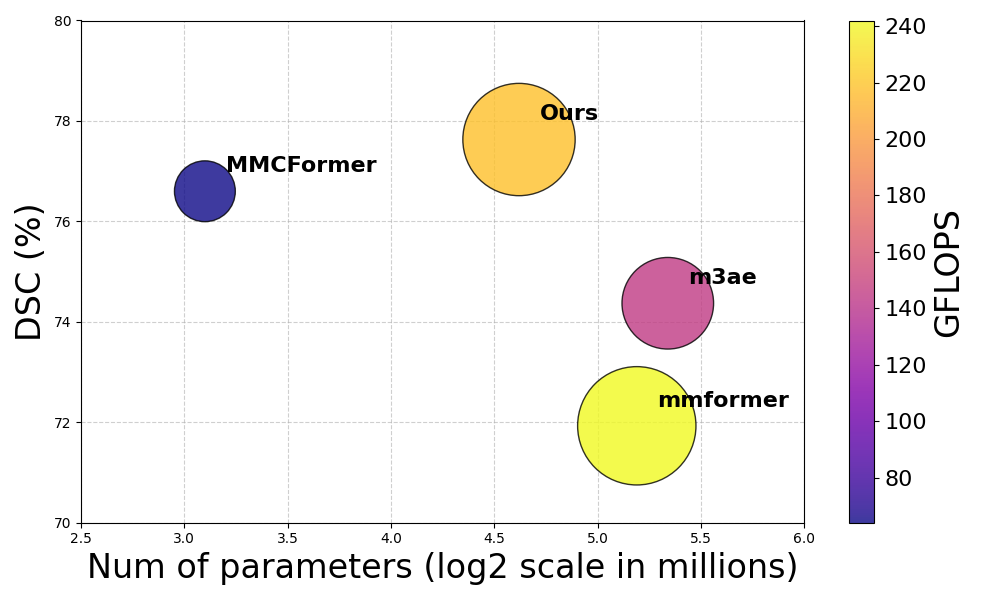}
    \caption{Deployment model size (in log2 scale) and mean Dice similarity scores (DSCs) across all missing-modal situations on BraTS 2018 test split; circle size indicates GFLOPS. }
    \label{fig:f4}
\end{figure}

Moreover, as shown in Fig.~\ref{fig:f4}, our method achieves a favorable trade-off between model efficiency and segmentation performance. It operates with significantly fewer parameters than existing state-of-the-art methods, while maintaining superior accuracy. This efficiency is especially valuable in real-world scenarios with limited computational resources. Although MMCFormer is lightweight per model, it requires training a separate model for each missing modality setting, resulting in exponential parameter growth as the number of modalities increases. In contrast, our method supports arbitrary modality combinations within a single unified model, offering both robustness and scalability.

To evaluate the generalizability of our method, we conducted experiments on the BraTS 2020 dataset and compared it with state-of-the-art methods including U-HVED, mmFormer, M\textsuperscript{3}AE, and LS3M~\cite{cvpr:1}. As shown in Tab.~\ref{tab:brats20}, our method consistently outperforms all baselines across the three tumor subregions, achieving average Dice improvements of 2.2\% (ET), 1.0\% (TC), and 0.2\% (WT) over the strongest baseline LS3M. It achieves the highest Dice scores in 11/15 ET, 10/15 TC, and 11/15 WT cases.
\begin{figure}
    \centering
    \includegraphics[width=\linewidth]{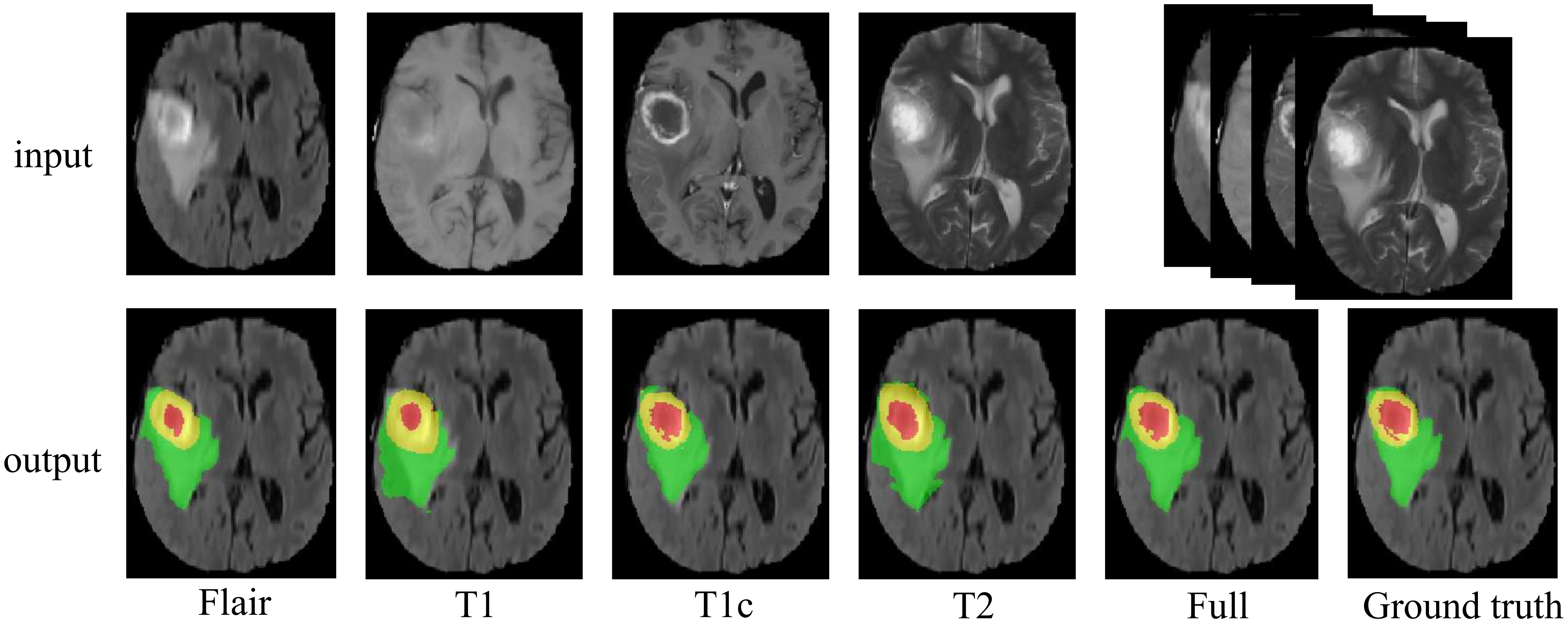}
    \caption{Example segmentation results of our method on BraTS2018, focusing on four extreme cases where only one single modality remains, as well as the case with all modalities available. Green: peritumoral edema; yellow: enhancing tumor; and red: necrotic and non-enhancing tumor core.}
    \label{fig:f5}
\end{figure}

Fig.~\ref{fig:f5} presents qualitative results under extreme conditions, where only a single modality is provided as input. These visualizations demonstrate that our model can still produce accurate and reliable segmentations, highlighting its robustness to missing modalities. In particular, results with only the T1c modality show clearly superior delineation of the enhancing tumor region, reaffirming the importance of T1c in capturing tumor enhancement.

\begin{figure}
    \centering
    \includegraphics[width=\linewidth]{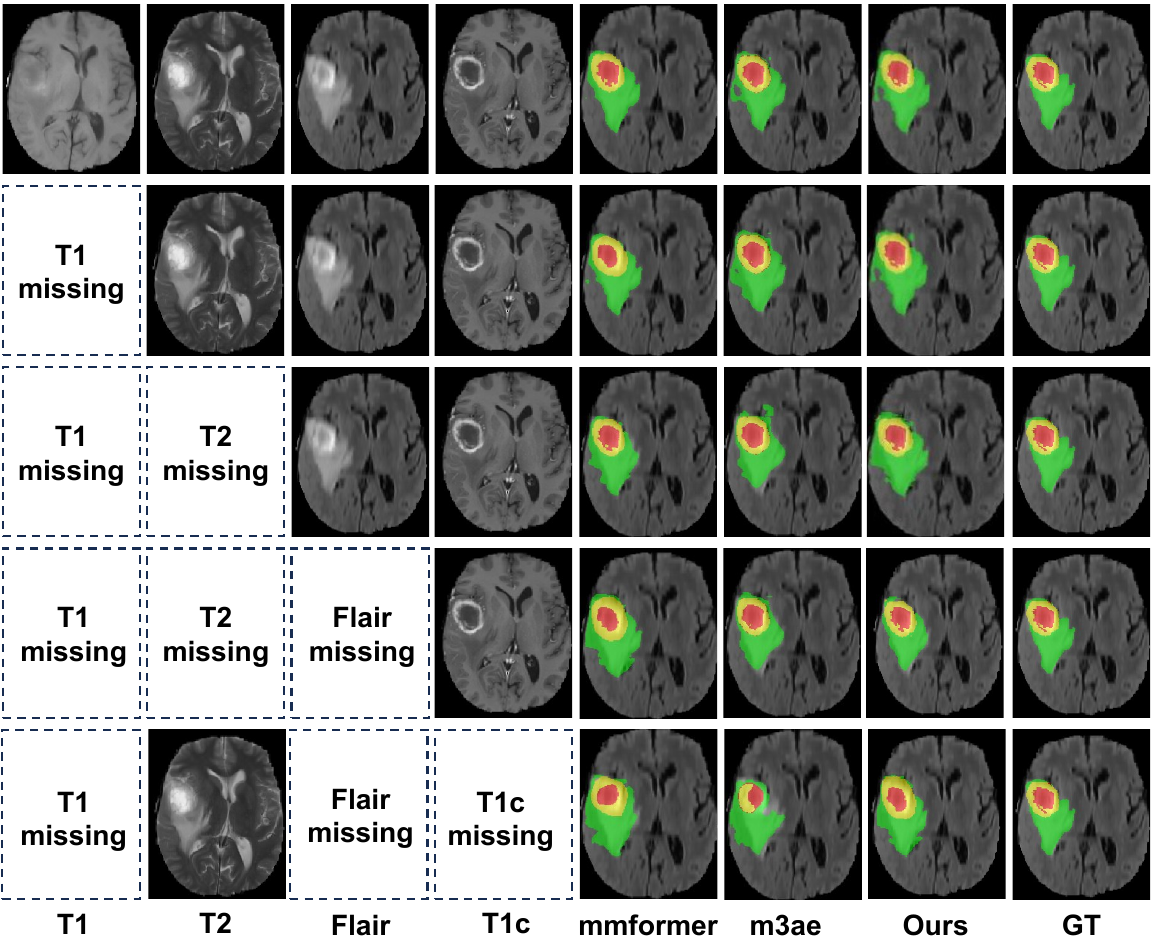}
    \caption{Qualitative segmentation results of different methods, including mmFormer, M3AE, and our approach, under various missing-modality settings. Green: peritumoral edema; yellow: enhancing tumor; and red: necrotic and non-enhancing tumor core.}
    \label{fig:f6}
\end{figure}

Figure~\ref{fig:f6} presents qualitative comparisons between our method and two state-of-the-art approaches, mmFormer \cite{c:7} and M3AE \cite{c:12}, under different modality combinations. Consistent with the quantitative results, all methods achieve reasonable segmentation quality when the most informative modality, T1c, is available, even in the single-modality setting. However, when only T2 is available, both mmFormer and M3AE show clear performance degradation, particularly in the red and yellow regions, with evident spatial misalignment. In contrast, our method remains more robust and produces more coherent delineations across all tumor subregions. These results demonstrate that the proposed virtual nodes guided dynamic graph effectively preserves structural information under missing-modality conditions, highlighting its advantage in extreme missing-modality scenarios.

\subsection{Ablation Study} To evaluate the contribution of each component in our framework, we conducted ablation studies summarized in Table~\ref{tab:ablation}. Specifically, we removed one component at a time—virtual nodes, dynamic edge connections, or the heterogeneous weight matrix—to assess its individual impact. For the variant without dynamic connections, we adopted a fully connected static graph as a baseline. For the variant without the heterogeneous matrix, we used a standard Graph Attention Network (GAT) formulation.

The results show that removing any single component degrades performance, with the most notable drop observed in enhancing tumor segmentation. This is consistent with clinical observations, as ET segmentation heavily relies on the presence of T1c. The tumor core and whole tumor regions are comparatively less sensitive, showing only moderate declines. Notably, virtual nodes play a key role in compensating for missing modalities, and their removal causes the largest performance degradation, particularly for the ET region.

\begin{table}[t]
    \centering
    \begin{tabular}{c|c|ccc}
    \hline
        & ablation method & ET & TC  & WT \\ \hline
    w/o & virtual nodes & 62.3 & 79.4 & 86.2 \\
    w/o & dynamic connection & 64.6 & 80.5 & 86.3 \\ 
    w/o & heterogeneous matrix& 62.9 & 80.9 & 86.3 \\  \hline
        & total method& \textbf{66.0} & \textbf{81.4} & \textbf{86.7} \\
    \hline
    \end{tabular}
    \caption{Ablation study on BraTS2018. The mean performance (DSC, \%) over all modality combinations is reported.}
    \label{tab:ablation}
\end{table}

Table~\ref{tab:length} examines the impact of virtual node length. While intuition might suggest that a longer virtual node enhances performance by enabling richer feature representation from missing modalities, our results show optimal performance at a length of 32—beyond which performance declines significantly. This counterintuitive finding indicates inherent limits to the modality-invariant information inferable from missing inputs: a moderate length (32) suffices to capture essential features, whereas larger dimensions introduce redundant or noisy information that impairs generalization. Additionally, increasing the virtual node length elevates computational overhead in memory and inference time. Thus, we select a length of 32, as it balances representational capacity and computational cost to achieve the best trade-off between effectiveness and efficiency.
\begin{table}[t]
    \centering
    \begin{tabular}{c|ccc|c}
    \hline
  length & ET   & TC   & WT & Avg\\ \hline
       0 & 62.3 & 79.4 & 86.2 & 75.9\\
      16 & 65.4 & 80.9 & 86.6 & 77.6\\ 
      32 & \textbf{66.0} & \textbf{81.4} & \textbf{86.7}  & \textbf{78.0}\\  
      64 & 63.3 & 79.6 & 86.6 & 76.5\\
     128 & 63.1 & 78.4 & 86.5 & 76.0\\
    \hline
    \end{tabular}
    \caption{Different length of virtual nodes of our model experimented on BraTS2018. The mean performance (DSC, \%) over all modality combinations is reported.}
    \label{tab:length}
\end{table}
\section{Conclusion}
In this work, we present a graph-based framework for multimodal brain tumor segmentation that effectively addresses missing-modality scenarios. By introducing virtual nodes and an adaptive connectivity mechanism, our method accommodates arbitrary modality combinations within a unified one-stage training paradigm and avoids the inefficiency of conventional two-stage pipelines. Extensive experiments on BraTS2018 and BraTS2020 show that our approach achieves competitive or superior performance compared with state-of-the-art methods. In future work, we plan to extend the framework to broader medical segmentation tasks and explore its potential as a plug-and-play solution for missing-modality challenges beyond the medical domain.

\section*{Acknowledgements}
This work was supported by the National Key Research and Development Program of China (Grant No. 2022ZD0118001), the National Natural Science Foundation of China (Grant Nos. 62332017, U22A2022), and was conducted in part at the MOE Key Laboratory of Advanced Materials and Devices for Post-Moore Chips, the Beijing Key Laboratory of Big Data Innovation and Application for Skeletal Health Medical Care, and the Beijing Advanced Innovation Center for Materials Genome Engineering.